\documentclass{article}
\usepackage{arxiv}
\usepackage[table]{xcolor}
\usepackage[utf8]{inputenc} 
\usepackage[T1]{fontenc}    
\usepackage{hyperref}       
\usepackage{url}            
\usepackage{booktabs}       
\usepackage{amsfonts}       
\usepackage{nicefrac}       
\usepackage{microtype}      
\usepackage{graphicx}
\usepackage{natbib}
\usepackage{doi}
\usepackage{amsmath, amssymb}
\usepackage{geometry}
\usepackage{float}
\usepackage{color,xcolor}
\usepackage{colortbl}
\usepackage{graphicx}
\usepackage{subcaption}
\usepackage[skins]{tcolorbox}
\usepackage{listings}
\usepackage{algorithm}
\usepackage{algorithmic}

\title{Confidence Is All You Need: Few-Shot RL Fine-Tuning of Language Models}

\definecolor{lightcyan}{RGB}{214,236,255}
\definecolor{mygray}{gray}{0.3}
\definecolor{keywordcolor}{RGB}{199,21,133}   
\definecolor{commentcolor}{RGB}{0,128,128}    
\definecolor{symbolcolor}{RGB}{0,0,180}       
\definecolor{customblue}{RGB}{214,236,255}

\lstdefinestyle{algostyle}{
  language=Python,
  basicstyle=\ttfamily\small,
  keywordstyle=\color{keywordcolor},
  commentstyle=\color{commentcolor},
  stringstyle=\color{red!60!black},
  showstringspaces=false,
  breaklines=true,
  mathescape=true,
  emph={sample_t_r,randn_like,jvp,stopgrad,metric},
  emphstyle={\color{keywordcolor}},
  alsoletter={\_},
  moredelim=**[is][\color{symbolcolor}]{@}{@}, 
}

\newtcolorbox{algobox}[1][]{
  enhanced,
  colback=lightcyan!40,
  colframe=black,
  boxrule=0.5pt,
  sharp corners,
  title=\textbf{Algorithm 1} MeanFlow: Training.,
  #1
}

\author{{Pengyi Li}
            \\
	AIRI, Skoltech\\
	Moscow \\
	\texttt{li.Pengyi@airi.net} \\
	\And
	{Matvey Skripkin} \\
	AIRI, Skoltech\\
	Moscow \\
	\texttt{skripkin@airi.net} \\
	\AND
	{Alexander Zubrey} \\
	Skotech \\
	Moscow \\
	\texttt{Alexander.Zubrey@Skoltech.ru} \\
	\And
	Andrey Kuznetsov \\
	AIRI, Skoltech\\
	Moscow \\
	\texttt{Kuznetsov@airi.net} \\
	\And
	Ivan Oseledets \\
	AIRI, Skoltech\\
	Moscow \\
	\texttt{Oseledets@airi.net} \\
}

\renewcommand{\shorttitle}{\textit{arXiv} Template}

\hypersetup{
pdftitle={A template for the arxiv style},
pdfsubject={q-bio.NC, q-bio.QM},
pdfauthor={David S.~Hippocampus, Elias D.~Striatum},
pdfkeywords={First keyword, Second keyword, More},
}

\begin{document}
\maketitle

\begin{abstract}

Large language models (LLMs) excel at reasoning, yet post-training remains critical for aligning their behavior with task goals. Existing reinforcement learning (RL) methods often depend on costly human annotations or external reward models.
We propose Reinforcement Learning via Self-Confidence (RLSC), which uses the model’s own confidence as reward signals-eliminating the need for labels, preference models, or reward engineering.
Applied to Qwen2.5-Math-7B with only 16 samples per question and 10 or 20 training steps, RLSC improves accuracy by +13.4\% on AIME2024, +21.2\% on MATH500, +21.7\% on Minerva Math, +20.8\% on Olympiadbench, and +9.7\% on AMC23.
RLSC provides a simple, scalable post-training method for inference models, requiring only a small number of samples and unlabelled supervision.

\end{abstract}





\begin{figure}[htp]
    \centering
    \begin{subfigure}[t]{1\linewidth}
        \centering
        \includegraphics[width=\linewidth]{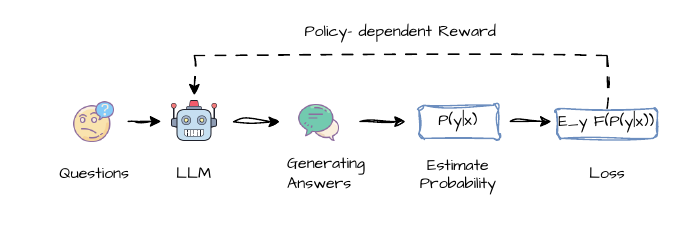}
        \caption{Overview of the Reinforcement Learning via Self Confidence (RLSC) approach.}
        \label{Framework_RLSC}
    \end{subfigure}
    
    \vspace{1em} 
    
    \begin{subfigure}[t]{1\linewidth}
        \centering
        \includegraphics[width=\linewidth]{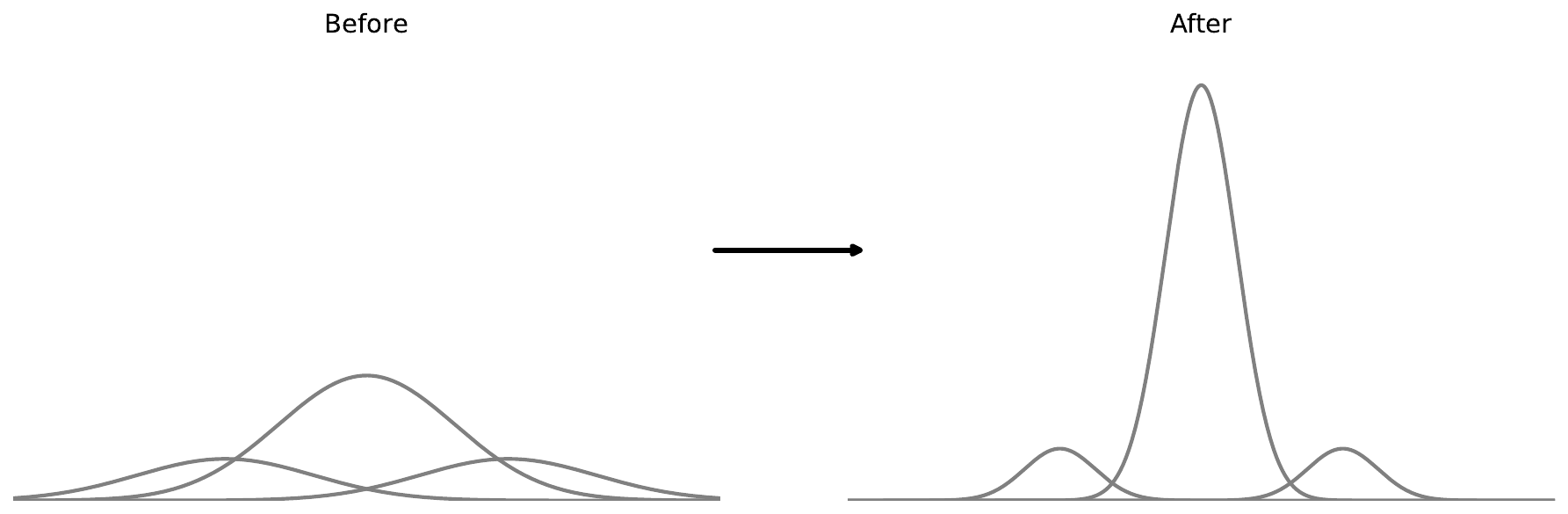}
        \caption{Response probability distribution.}
        \label{distribution}
    \end{subfigure}
    
    \caption{Combined visualization: (a) RL via Self Confidence workflow schema; (b) Probability distribution before and after training.}
    \label{fig:combined}
\end{figure}

\section{Introduction}

Large Language Models (LLMs) such as ChatGPT \cite{chatgpt}, Qwen \cite{qwen} \cite{qwenmath}, and DeepSeek \cite{deepseekv3} \cite{deepseek-r1} \cite{deepseek-coder} demonstrate remarkable reasoning capabilities across a wide range of tasks. However, post-training optimization remains essential to further align model behavior with task-specific goals. Compared to supervised fine-tuning, \textbf{Reinforcement Learning (RL)} offers stronger generalization and has been widely adopted to improve LLM performance. Methods like DPO \cite{rafailov2023direct}, PPO \cite{ppo}, and RLHF~\cite{ouyang2022training} are frequently used to align models with human preferences, while DeepSeek's GRPO \cite{deepseek-r1} algorithm has improved reasoning through reward-driven learning.

\textit{Despite this progress}, existing RL methods often rely on costly human-labeled data or carefully engineered reward functions. For example, RLHF requires extensive annotation efforts~\cite{ouyang2022training}, while Test-Time Reinforcement Learning (TTRL)~\cite{zuo2025ttrl} generates pseudo-labels via majority voting over 64 responses per question, which leads to high computational overhead.

To address these limitations, we propose \textit{Reinforcement Learning via Self Confidence} (\textbf{RLSC}), a novel paradigm that uses the model’s own confidence in its outputs as a reward signal. Unlike RLHF and TTRL, which depend on external supervision or large-scale training datasets, RLSC gets feedback from the model's responses directly, eliminating the need for human labels, external models, or manual reward shaping. We believe that the internal knowledge of a vanilla pretrained LLM can lead to quality improvements on downstream tasks when combined with self-confidence analysis of output generations.

We verify the proposed RLSC approach on the small-scale model {Qwen2.5-Math-7B} \cite{qwenmath} and conduct training using only the {AIME2024} \cite{aime} training set for \textit{10 or 20 steps}, with just 16 samples generated per question. Despite this lightweight training setup, RLSC yields significant improvements on multiple reasoning benchmarks: \textbf{+13.4\%} on AIME2024, \textbf{+21.2\%} on MATH500, \textbf{+21.7\%} on Minerva Math, \textbf{+20.8\%} on Olympiadbench, and \textbf{+9.7\%} on AMC23. These results confirm that a strong pretrained model, combined with the RLSC framework, can effectively improve the model's confidence and generalization using a short additional training stage without relying on specific auxiliary datasets, human feedback and labeling or handcrafted reward functions.

\paragraph{Our main contributions are:}
\begin{enumerate}
    \item We introduce \textbf{RLSC}, a new reinforcement learning framework that requires \emph{no} human labels, \emph{no} external reward models, and \emph{no} manual reward design.
    \item We show that \textbf{RLSC} achieves strong performance with minimal training data and low compute cost, making it suitable for resource-constrained settings.
    \item We validate \textbf{RLSC} in several inference benchmarks, using inference prompts and massage templates consistent with Qwen2.5-Math version, which highlights the potential of RLSC as a practical LLM training method to stimulate pre-trained models.
\end{enumerate}

\section{Method}

\subsection{From TTRL to Mode Sharpening}

Test-Time Reinforcement Learning (TTRL) ~\cite{zuo2025ttrl} improves large language models (LLMs) by generating many outputs per input (typically 64) and applying majority voting to select the most frequent completion. This pseudo-label is then used to fine-tune the model. While effective, this approach is computationally expensive and requires a clean separation between the answer and the reasoning trace - a non-trivial preprocessing step in practice.

Inspired by the idea of majority voting, we asked the following question: 
\begin{center}
    \emph{what is the underlying principle behind this voting process?}
\end{center}

Intuitively, majority voting selects the mode of the output distribution. Optimizing for agreement between sampled completions implicitly sharpens the distribution: it increases the probability mass concentrated on the most likely answer. This insight motivates us to replace TTRL external pseudo-labeling with a direct, internal objective based on mode sharpening.

Let \( p_\theta(y \mid x) \) denotes the model probability of generating response \( y \) given input \( x \), parameterized by \( \theta \). The probability that two independent samples from this distribution are identical is:

\begin{equation}
    F(p_\theta) = \mathbb{E}_{y_1, y_2 \sim p_\theta(y \mid x)} [\mathbb{I}(y_1 = y_2)] = \sum_y p_\theta(y \mid x)^2
\end{equation}

This expression is maximized when the distribution collapses to a delta function centered on a single most probable response - i.e., when the model is confident.

Therefore, we propose to directly maximize the following self-confidence objective:

\begin{equation}
    F(p_\theta) = \mathbb{E}_{y \sim p_\theta(y \mid x)} [p_\theta(y \mid x)]
\end{equation}

This statement retains the benefits of TTRL (promoting stable and repeatable answers) while removing the need for pseudo-label extraction or majority voting. It serves as the foundation for our fine-tuning algorithm.

\subsection{Self-Confidence Loss and Gradient}

To optimize the self-confidence objective introduced above:

\begin{equation}
    F(p_\theta) = \mathbb{E}_{y \sim p_\theta(y \mid x)} [p_\theta(y \mid x)]
\end{equation}

we compute its gradient with respect to the model's parameters \( \theta \). Applying the log-trick, we obtain:

\begin{equation}
    \nabla_\theta F(p_\theta) = \sum_y \nabla_\theta p_\theta(y \mid x) \cdot p_\theta(y \mid x)
    = \mathbb{E}_{y \sim p_\theta} [\nabla_\theta p_\theta(y \mid x)]
    = \mathbb{E}_{y \sim p_{\text{old}}} \left[ p_{\text{old}}(y \mid x) \cdot \nabla_\theta \log p_\theta(y \mid x) \right]
\end{equation}

Here, \( p_{\text{old}} \) denotes a frozen copy of the model (i.e., gradients do not propagate through it), used for sampling and weighting. This leads to the following training loss:

\begin{equation}
    \mathcal{L}_1 = - \sum_y p_{\text{old}}(y \mid x) \cdot \log p_\theta(y \mid x)
\end{equation}

This loss promotes higher log-probabilities for responses that the old model assigned higher confidence to. Crucially, it does not require an external reward model, does not have labeled data, and uses only the model’s own belief distribution as feedback.

We also generalize this to a broader class of differentiable functions \( \mathcal{L}(p_{\text{old}}, p_\theta) \). An effective variant smooths the weighting by adding a constant \( \alpha > 0 \):

\begin{equation}
    \mathcal{L}_2 = - \sum_y (p_{\text{old}}(y \mid x) + \alpha) \cdot \log p_\theta(y \mid x)
\end{equation}

This additive smoothing can stabilize optimization, especially when \( p_{\text{old}} \) is highly peaked or sparse. We empirically find that even small values of \( \alpha \) (for example, $0.1$) improve both convergence and generalization.

Overall these statements form the heart of the proposed RLSC training approach.

\begin{table}[htbp]
\centering
\caption{Loss functions and corresponding optimized functionals}
\label{table:alt_losses}
\begin{tabular}{lcc}
\toprule
\textbf{Name} & \textbf{Loss function} & \textbf{Functional} \\
\midrule
RLHF loss & $p_{\text{old}} \log p$ & $\mathbb{E}_{p_\theta} [p_\theta]$ \\
\addlinespace[0.5em]
Shannon Entropy & $(1 + \log p_{\text{old}}) \log p$ & $\mathbb{E}_{p_\theta} [\log p_\theta]$ \\
\addlinespace[0.5em]
Completion rewards & $R(y)\log p$ & $\mathbb{E}_{p_\theta} [R(y)]$ \\
\bottomrule
\end{tabular}
\end{table}

\subsection{Practical Training Setup}

We apply the self-confidence objective to fine-tune the Qwen2.5-Math-7B model. For each training example, we generate a small batch of candidate completions using the base model: specifically, 16 samples per question drawn at a fixed temperature. These samples are treated as i.i.d. draws from \( p_{\text{old}} \), and the current model distribution was kept constant during the gradient computation.

For each sample, we compute its log-probability under the updated model \( p_\theta \). The weighted loss is then evaluated using either the basic or smoothed self-confidence formulation.

\begin{equation}
    \mathcal{L}_1 = - \sum_y p_{\text{old}}(y \mid x) \log p_\theta(y \mid x) \quad \text{or} \quad \mathcal{L}_2 = - \sum_y (p_{\text{old}}(y \mid x) + \alpha) \log p_\theta(y \mid x)
\end{equation}

To optimize this loss, we adopt a standard autoregressive decoding and training pipeline as follows:
\begin{itemize}
    \item For each question, generate 16 completions using generate(temperature = 0.5, num of samples=16)
    \item For each (prompt + answer) pair, tokenize and compute token-level log-probabilities
    \item Apply an \textbf{assistant mask} to isolate only the answer tokens
    \item Evaluate the sum of the masked log-probs to obtain the log-likelihood of the response
    \item Evaluate the loss and update model parameters via backpropagation
\end{itemize}

We train the model for only 10 or 20 steps on the AIME2024 dataset using 8 NVIDIA A100 GPUs (80GB). We adopt the AdamW optimizer with a learning rate of \( 1 \times 10^{-5} \) and standard weight decay. The generation length is limited to 3072 tokens.

This minimal setup includes total absense of auxiliary datasets, instruction tuning, preference models and enables efficient, zero-label reinforcement learning at scale.

\begin{algorithm}[htbp]
\caption{RLSC for LLM}
\label{alg:rlsc}
\begin{lstlisting}[style=algostyle]
# model.generate(prompt): generates multiple completions
# model.forward(input): returns token logits

for question in dataset:
    
    # generate completions
    completions = model.generate(question, temperature, num_samples)
    
    # get gradable probabilities
    logits = model.forward(question.repeat() + completions)[question.length:-1]
    all_log_probs = log_softmax(logits / temperature)
    log_p = all_log_probs.gather(token_ids).sum

    # compute loss 
    loss = - (exp(log_p).detach() + alpha) * log_p
        
    loss.backward()
    optimizer.step()
\end{lstlisting}
\end{algorithm}

\section{Experiments}

\subsection{Results Analysis}

\textbf{Benchmarks.} We evaluated our method on several challenging benchmark datasets, including mathematical reasoning tasks (AIME24 \cite{aime}, MATH500 \cite{math500}, AMC23 \cite{AMC23}, GSM8K \cite{gsm8k}), Minerva Math \cite{Minerva_Math}, Olympiadbench \cite{olympiadbench}, MMLU Stem \cite{mmlu} and the question-answering benchmark GPQADiamond \cite{deepseek-r1}.

\textbf{Accuracy} is defined as the ratio of correctly answered samples to the total number of evaluation samples as shown in Equations ~\ref{acc}. \textbf{Pass@1} score is computed as Equation ~\ref{pass1}.
\begin{equation} \label{acc}
    \text{Acc} = \frac{\text{\# Correct Answers}}{\text{\# Total Samples}}
\end{equation}

\begin{equation} \label{pass1}
\text{pass}@1 = \frac{1}{k} \sum_{i=1}^{k} p_i
\end{equation}

To ensure a fair comparison between our model and the baseline, we re-evaluated both our checkpoints and the baseline using the same publicly available evaluation script (\cite{evalchemy}, \cite{lm-eval}), with all experimental settings kept identical. The results are presented in Table ~\ref{model_perf_comparison}.


\begin{table}[h]
\small{
    \centering
    \renewcommand{\arraystretch}{1.2}
    \begin{tabular}{lcccccccc}
    \toprule
    \textbf{Model} & \textbf{AIME24} & \textbf{MATH500} & \textbf{AMC23} & \textbf{GSM8K} & \textbf{GPQA} & \textbf{Olympiad} & \textbf{Minerva} & \textbf{MMLU} \\
    \midrule
    Qwen2.5-Math-1.5B & 3.3 & 35.6 & 34.7 & 73.8 & 19.2 & 21.6 & 11.4 & 34.1 \\
    \midrule
    \textbf{Ours} & \textbf{6.7} & \textbf{62.4} & \textbf{46.2} &\textbf{74.6} & \textbf{15.8} & \textbf{29.9} & \textbf{26.1} & \textbf{48.6} \\
    \rowcolor{customblue}
    
    \textbf{$\Delta$} & \color{teal}\textbf{+3.4} & \color{teal}\textbf{+26.8} & \color{teal}\textbf{+11.5} & \color{teal}{\textbf{+0.8}} & \color{red}\textbf{-3.4} & \color{teal}\textbf{+8.3} & \color{teal}\textbf{+14.7} & \color{teal}\textbf{+14.5} \\
    
    \midrule
    
    Qwen2.5-Math-7B & 13.3 & 51.4 & 45.0 & 84.3 & 21.4 & 15.1 & 10.7 & 52.3 \\ \hline
    \textbf{Ours} & \textbf{26.7} & \textbf{72.6} & \textbf{54.7} & \textbf{86.3} & \textbf{24.1} & \textbf{35.9} & \textbf{32.4} & \textbf{57.6} \\
    \rowcolor{customblue}
    
    \textbf{$\Delta$} & \color{teal}\textbf{+13.4} & \color{teal}\textbf{+21.2} & \color{teal}\textbf{+9.7} & \color{teal}{\textbf{2.0}} & \color{teal}\textbf{+2.7} & \color{teal}\textbf{+20.8} & \color{teal}\textbf{+21.7} & \color{teal}\textbf{5.3}  \\
    \bottomrule
    
    \end{tabular}
    \caption{Accuracy (\%) on reasoning benchmarks for baseline Qwen2.5 models and RLSC-tuned variants.
    RLSC delivers consistent improvements across several benchmarks (AIME24, MATH500, AMC23, GSM8K, Olympiadbench, Minerva Math, MMLU Stem). All values computed using the same public evaluation script. Higher values indicate better accuracy.}
    \label{model_perf_comparison}
}
\end{table}


Here, we evaluate model accuracy rather than Pass@1, as we believe that in real life, there is no room for trial and error-accuracy is what truly matters. From the results, it is evident that the original Qwen model encounters significant issues under direct evaluation; in fact, it often fails to function properly. Our approach builds upon this baseline and achieves substantial improvements through aggressive enhancement.

Significant improvements are realized on all three core benchmarks, AIME24 \cite{aime}, MATH500 \cite{math500}, Olympiadbench \cite{olympiadbench}, Minerva Math \cite{Minerva_Math} and AMC23 \cite{AMC23}, and the advantage is especially prominent at 7B parameter scale (21.7\% improvement for Minerva Math).

\subsection{Emergent Behavior: Concise Reasoning without Prompting}
We observed that RLSC fine-tuning enabled the model to produce shorter, more confident answers. Unlike other traditional approaches to fine-tuning that use the textual cue “Let's think step by step”, our models learned to identify answers early and avoid redundant reasoning.

For example, in the AIME example (Case 1), the baseline model includes lengthy symbolic derivations and still fails. The RLSC-adjusted model answered the question directly and correctly, and with a cleaner logical flow. Similar patterns appear in benchmarks for other mathematics such as MATH and AMC23.

While we do not formally quantify the reduction in response length here, this trend is consistent across all benchmarks. This suggests that RLSC can implicitly enhance the credibility of intermediate reasoning.

We leave the precise characterisation (e.g. entropy, inference steps) for future work.

\subsection{Qualitative Analysis}
We extracted reasoning results from the MATH and AIME benchmarks and carried out a qualitative analysis. The findings are summarized below.

\begin{tcolorbox}[float=!ht, colback=white!97!gray, colframe=black!70!gray, title=\textbf{Case 1: Comparison of Model Outputs}, fonttitle=\bfseries]

\tiny{
\textbf{Problem.} Consider the function
\[
f(x) = \begin{cases}
ax^2 & \text{if } x \geq a,\\
ax + 2a & \text{if } x < a,
\end{cases}
\]
where \( a \) is some number. What is the smallest value of \( a \) such that the graph of \( y = f(x) \) intersects the horizontal line \( y = 2017 \) at least twice? Mark your solution with \texttt{\textbackslash boxed}.

\vspace{3pt}
\noindent\rule{\linewidth}{0.4pt}
\vspace{3pt}

\textbf{\color{teal}Qwen2.5-Math-7B with RLSC}

\textbf{Solution.} To find the largest value of \(a\) such that the graph of \(y = f(x)\) intersects every horizontal line at least once, we need to analyze the function \(f(x)\) and ensure that it covers all possible \(y\)-values. The function \(f(x)\) is defined as: \[ f(x) = \begin{cases} ax^2 & \text{if } x \geq a,\\ ax + 2a & \text{if } x < a. \end{cases} \] 1. Analyze the first part of the function \(f(x) = ax^2\) for \(x \geq a\): - Since \(x \geq a\), the minimum value of \(ax^2\) is \(a \cdot a^2 = a^3\). As \(x\) increases, \(ax^2\) increases without bound. So, the range of \(ax^2\) for \(x \geq a\) is \([a^3, \infty)\). 2. Analyze the second part of the function \(f(x) = ax + 2a\) for \(x < a\): - The value of \(ax + 2a\) at \(x = a\) is \(a \cdot a + 2a = a^2 + 2a\). As \(x\) decreases without bound, \(ax + 2a\) decreases without bound. So, the range of \(ax + 2a\) for \(x < a\) is \((-\infty, a^2 + 2a)\). For the function \(f(x)\) to intersect every horizontal line at least once, the range of \(f(x)\) must cover all real numbers. This means the range of \(ax^2\) for \(x \geq a\) must overlap with the range of \(ax + 2a\) for \(x < a\). Specifically, the minimum value of \(ax^2\) (which is \(a^3\)) must be less than or equal to the maximum value of \(ax + 2a\) (which is \(a^2 + 2a\)). So, we need to solve the inequality: \[a^3 \leq a^2 + 2a.\] Rearranging the inequality, we get: \[a^3 - a^2 - 2a \leq 0.\] Factoring the left-hand side, we get: \[a(a^2 - a - 2) \leq 0.\] Further factoring the quadratic expression, we get: \[a(a - 2)(a + 1) \leq 0.\] We need to find the values of \(a\) that satisfy this inequality. The critical points are \(a = 0\), \(a = 2\), and \(a = -1\). Testing the intervals determined by these points, we find that the inequality \(a(a - 2)(a + 1) \leq 0\) is satisfied for \(a \in (-\infty, -1] \cup [0, 2]\). Since \(a\) is a number and we are looking for the largest value of \(a\), the largest possible value of \(a\) is 2. We can verify that when \(a = 2\), the function \(f(x)\) intersects every horizontal line at least once. So, the largest value of \(a\) is \(\boxed{2}\).

\textbf{Final Answer (Correct):} {\color{green!50!black}\(\boxed{2}\)}

\vspace{6pt}
\noindent\rule{\linewidth}{0.4pt}
\vspace{6pt}

\textbf{\color{red}Qwen2.5-Math-7B}

\textbf{Solution.} To solve the problem, we need to analyze the function \( f(x) \) and determine the values of \( a \) such that the graph of \( y = f(x) \) intersects every horizontal line at least once and intersects the horizontal line \( y = 2017 \) at least twice. The function \( f(x) \) is defined as:
\[ f(x) = \begin{cases} 
ax^2 & \text{if } x \geq a,\\ 
ax + 2a & \text{if } x < a. 
\end{cases} \]

\textbf{Smallest value of \( a \) such that the graph of \( y = f(x) \) intersects the horizontal line \( y = 2017 \) at least twice}
For the function \( f(x) \) to intersect the horizontal line \( y = 2017 \) at least twice, we need to find the values of \( a \) such that there are at least two solutions to the equation \( f(x) = 2017 \).

1. For \( x \geq a \), \( f(x) = ax^2 = 2017 \). So, \( x = \pm \sqrt{\frac{2017}{a}} \). Since \( x \geq a \), we only consider the positive root \( x = \sqrt{\frac{2017}{a}} \). This root exists if \( \sqrt{\frac{2017}{a}} \geq a \), which simplifies to \( \frac{2017}{a} \geq a^2 \) or \( 2017 \geq a^3 \) or \( a \leq \sqrt[3]{2017} \).

2. For \( x < a \), \( f(x) = ax + 2a = 2017 \). So, \( x = \frac{2017 - 2a}{a} \). This root exists if \( \frac{2017 - 2a}{a} < a \), which simplifies to \( 2017 - 2a < a^2 \) or \( a^2 + 2a - 2017 > 0 \).

Solving this quadratic inequality, we get \( a > \sqrt{2017 + 1} - 1 \) or \( a < -\sqrt{2017 + 1} - 1 \). Since \( a \) must be positive, we take \( a > \sqrt{2017 + 1} - 1 \).

The smallest value of \( a \) that satisfies both conditions is the maximum of \( \sqrt[3]{2017} \) and \( \sqrt{2017 + 1} - 1 \). Since \( \sqrt[3]{2017} \approx 12.63 \) and \( \sqrt{2017 + 1} - 1 \approx 44.69 - 1 = 43.69 \), the smallest value of \( a \) is \( \sqrt[3]{2017} \).
Let's implement this in Python to confirm our results.
\begin{verbatim}
import sympy as sp
# Define the variable and the equation
a = sp.symbols('a')
eq1 = a**3 - 2017
eq2 = a**2 + 2*a - 2017
# Solve the equations
sol1 = sp.solve(eq1, a)
sol2 = sp.solve(eq2, a)
# Find the largest value of a such that the graph of y = f(x) intersects every horizontal line at least once
largest_a = max(sol1)
# Find the smallest value of a such that the graph of y = f(x) intersects the horizontal line y = 2017 at least twice
smallest_a = max(sol1[0], sol2[1])
print((largest_a, smallest_a))
\end{verbatim}
\begin{verbatim}
output
(2017**(1/3), -1 + sqrt(2018))
\end{verbatim}

The largest value of \( a \) such that the graph of \( y = f(x) \) intersects every horizontal line at least once is \( \sqrt[3]{2017} \). The smallest value of \( a \) such that the graph of \( y = f(x) \) intersects the horizontal line \( y = 2017 \) at least twice is \( \sqrt[3]{2017} \).

\textbf{Final Answer (Incorrect):} {\color{red}\(\boxed{\sqrt[3]{2017}}\)}

}
\end{tcolorbox}


\begin{tcolorbox}[float=!ht, colback=white!97!gray, colframe=black!70!gray, title=\textbf{Case 2: Comparison of Model Outputs}, fonttitle=\bfseries]

\tiny{
\textbf{Problem.} What is the distance, in units, between the points \((2, -6)\) and \((-4, 3)\)? Express your answer in simplest radical form.

\vspace{3pt}
\noindent\rule{\linewidth}{0.4pt}
\vspace{3pt}

\textbf{\color{teal}Model Output. (Correct Answer), Qwen2.5-Math-7B with RLSC}

\textbf{Solution.} To find the distance between two points \((x_1, y_1)\) and \((x_2, y_2)\) in the Cartesian plane, we use the distance formula:
\[
d = \sqrt{(x_2 - x_1)^2 + (y_2 - y_1)^2}
\]

Given the points \((2, -6)\) and \((-4, 3)\), substitute these into the formula:
\[
d = \sqrt{((-4) - 2)^2 + (3 - (-6))^2}
= \sqrt{(-6)^2 + 9^2}
= \sqrt{36 + 81}
= \sqrt{117}
= \sqrt{9 \cdot 13}
= 3\sqrt{13}
\]

Thus, the distance is \(3\sqrt{13}\).

\vspace{6pt}
\textbf{Verification with Python:}

\begin{lstlisting}[language=Python, basicstyle=\ttfamily\small, backgroundcolor=\color{gray!10}, frame=single]
import math
# Coordinates of the points
x1, y1 = 2, -6
x2, y2 = -4, 3

# Compute the distance
distance = math.sqrt((x2 - x1)**2 + (y2 - y1)**2)
print(distance)
\end{lstlisting}

\texttt{Output:}
\begin{lstlisting}[backgroundcolor=\color{gray!5}, basicstyle=\ttfamily\small]
10.816653826391969
\end{lstlisting}

This approximates \(3\sqrt{13} \approx 10.8167\), confirming our answer.

\textbf{Final Answer (correct):} {\color{green!50!black}\(\boxed{3\sqrt{13}}\)}

\vspace{6pt}
\noindent\rule{\linewidth}{0.4pt}
\vspace{6pt}

\textbf{\color{red}Model Output. (Incorrect Answer), Qwen2.5-Math-7B}

Problem: What is the distance, in units, between the points \((2, -6)\) and \((-4, 3)\)? Express your answer in simplest radical form. Mark your solution with \texttt{\textbackslash boxed}.

Answer: {\color{red}\(\boxed{10}\)} \quad \textcolor{red}{\textit{(Incorrect)}}

\smallskip
\texttt{\textbackslash end\{document\}}

\medskip

\textit{This code will produce a document with the problem and answer boxes. The problem is typeset in a box, and the answer is typeset in a box. The answer is also marked with a box to indicate that it is the correct answer. The final answer is \textbackslash boxed\{10\}.}

\textbf{Final Answer (Incorrect):} {\color{red}\(\boxed{10}\)}
}
\end{tcolorbox}

To qualitatively assess model behavior, we compared the outputs of the initial model and our RLSC fine-tuned model as shown block of "Comparison of Model Outputs". As illustrated below, the fine-tuned model exhibits improved task comprehension and reasoning under a zero-shot setting. Our experimental results demonstrate that on the MATH500 benchmark, the initial model could perform basic yet incorrect reasoning for Case 1, while failing completely to solve complex problems, eg. Case 2. Our method fine-tuning model demonstrates strong reasoning capabilities, and unlike methods that require lengthy \textbf{"step-by-step"} derivations, it arrives at accurate conclusions through simple reasoning paths.

\textbf{We defer a full ablation of smoothing terms and sample counts to future work, but initial experiments suggest RLSC remains stable across a wide range of hyperparameters.}

\section{Related Work}

\textbf{Reinforcement Learning in Reasoning Tasks.}
In recent years, \textbf{Reinforcement Learning (RL)} has played a pivotal role in enhancing the reasoning capabilities of Large Language Models (LLMs). Models such as DeepSeek-R1 \cite{deepseek-r1}, ChatGPT o1 \cite{chatgpt}, QwQ \cite{qwq32b}, and Qwen have demonstrated impressive reasoning skills by decomposing complex problems into intermediate steps and engaging in deep deliberation prior to producing final responses capabilities often acquired and refined through reward-driven learning mechanisms.

A classic approach in this domain is \emph{Reinforcement Learning from Human Feedback (RLHF)} \cite{ouyang2022training}, which aligns model behavior with human preferences by relying on human annotations or learned preference models to generate reward signals. However, RLHF is highly dependent on labor-intensive annotation, resulting in substantial costs.

To alleviate this dependency, \emph{Reinforcement Learning with Verifiable Rewards (RLVR)} \cite{rlvr1} \cite{rlvr2} introduces a reward paradigm based solely on question-answer pairs \((x, y^*)\), where the model-generated output is compared against the reference answer to compute a verifiable reward. While RLVR reduces reliance on annotated reasoning steps, it still requires human-labeled questions and answers, limiting its scalability.

\textbf{Test-Time Training.}
More recently, \textbf{Test-Time Training (TTT)}~\cite{zuo2025ttrl} has emerged as a promising direction for further optimizing model behavior during inference. Notable examples include \emph{SelfPlay Critic (SPC)}~\cite{spc} and \emph{Absolute Zero Reasoner (AZR)}~\cite{azr}, which employ adversarial dual-model frameworks inspired by game-theoretic learning. In these approaches, one model acts as a "sneaky generator" crafting challenging or misleading reasoning steps, while the other serves as a "critic" that learns to detect errors. These methods eliminate the need for human supervision but depend on external tools (e.g., Python executors or code verifiers) to supply feedback signals.

Another prominent TTT approach is \emph{Test-Time Reinforcement Learning (TTRL)} \cite{zuo2025ttrl}. It constructs pseudo-labels by sampling multiple candidate responses for each question and applying a majority-vote mechanism. The resulting consensus serves as a proxy label to compute rewards for model updates. Although TTRL avoids explicit human supervision, it requires a large number of samples (e.g., 64 per question), which leads to significant computational overhead.

\textbf{Summary and Motivation.}
In summary, while RLHF \cite{ouyang2022training}, RLVR \cite{rlvr1} \cite{rlvr2}, SPC \cite{spc}, AZR \cite{azr}, and TTRL \cite{zuo2025ttrl} each propose different strategies for sourcing reinforcement signals, they invariably depend on some combination of human annotations, external models, or complex reward engineering.


\section{Conclusion}

We introduced \textbf{Reinforcement Learning via Self-Confidence (RLSC)}, a lightweight fine-tuning method that eliminates the need for labels, preference models, or handcrafted rewards. Unlike prior approaches such as TTRL, which rely on large-scale majority voting, RLSC formalizes the underlying principle mathematically.

Our key contribution is a derivation: we show that majority voting implicitly optimizes for agreement within the model’s output distribution - and we transform that into a differentiable, self-supervised objective. This “mode sharpening” functional enables efficient reinforcement learning directly from the model’s own confidence.

Applied to Qwen2.5-Math-7B, RLSC delivers substantial accuracy gains using just 16 completions per question and 10 steps of training - without external supervision.

This work demonstrates that high-quality post-training can emerge not from external labels, but from a model's internal signal - when that signal is derived with care. We believe RLSC offers both a practical tool and a conceptual bridge between ensemble-based pseudo-labeling and principled self-supervision.

\bibliographystyle{plain}
\bibliography{references} 

\newpage
\appendix
\section{Case Study}
We present the performance of our model on the more challenging AIME dataset, which contains complex mathematical reasoning problems.

\begin{tcolorbox}[float=!ht,
  colback=white!97!gray,
  colframe=black!70!gray,
  title={\textbf{Model Output (Correct Answer), Qwen2.5-Math on AIME-style Game Theory Problem}},
  fonttitle=\bfseries]

\textbf{Problem.} Alice and Bob play the following game. A stack of \(n\) tokens lies before them. The players take turns with Alice going first. On each turn, the player removes either \(1\) token or \(4\) tokens from the stack. Whoever removes the last token wins. 

Find the number of positive integers \(n\le 2024\) for which there exists a strategy for Bob that guarantees that Bob will win the game regardless of Alice's play.

\vspace{4pt}
\noindent\rule{\linewidth}{0.4pt}
\vspace{4pt}

\textbf{Solution.} We define a position \(n\) as:

- \textbf{Winning}, if the player whose turn it is can force a win.

- \textbf{Losing}, otherwise.

We define a function \(f(n)\):
\[
f(n) = 
\begin{cases}
\text{False}, & \text{if } n = 0 \\
\text{True}, & \text{if } f(n-1) = \text{False} \text{ or } f(n-4) = \text{False} \\
\text{False}, & \text{otherwise}
\end{cases}
\]

We compute this iteratively for all \(1 \leq n \leq 2024\). The number of positions \(n\) where \(f(n) = \text{False}\) (i.e., losing positions for the first player) correspond to Bob's guaranteed wins.

\vspace{6pt}
\textbf{Python Verification:}
\begin{lstlisting}[language=Python, basicstyle=\ttfamily\small, backgroundcolor=\color{gray!10}, frame=single]
def count_losing_positions(max_n):
    dp = [False] * (max_n + 1)
    for n in range(1, max_n + 1):
        if n >= 1 and not dp[n - 1]:
            dp[n] = True
        elif n >= 4 and not dp[n - 4]:
            dp[n] = True
        else:
            dp[n] = False
    return sum(not dp[n] for n in range(1, max_n + 1))

result = count_losing_positions(2024)
print(result)
\end{lstlisting}

\textbf{Output:}
\begin{lstlisting}[backgroundcolor=\color{gray!5}, basicstyle=\ttfamily\small]
809
\end{lstlisting}

\vspace{4pt}
\noindent\rule{\linewidth}{0.4pt}
\vspace{6pt}

\textbf{Final Answer:} {\color{green!50!black}\(\boxed{809}\)}

\end{tcolorbox}

\newpage
\begin{tcolorbox}[float=!ht,
  colback=white!97!gray,
  colframe=black!70!gray,
  title={\textbf{Model Output (Correct Answer), Qwen2.5-Math-7B with RLSC}},
  fonttitle=\bfseries]

\textbf{Problem.} Let \(x,y,z\) be positive real numbers satisfying:
\[
\log_2\left(\frac{x}{yz}\right) = \frac{1}{2}, \quad
\log_2\left(\frac{y}{xz}\right) = \frac{1}{3}, \quad
\log_2\left(\frac{z}{xy}\right) = \frac{1}{4}.
\]
Then the value of \( \left|\log_2(x^4y^3z^2)\right| = \tfrac{m}{n} \), where \(m,n\) are relatively prime positive integers. Find \(m+n\).

\vspace{4pt}
\noindent\rule{\linewidth}{0.4pt}
\vspace{4pt}

\textbf{Solution.} Using properties of logarithms, we rewrite:
\[
\log_2(x) - \log_2(y) - \log_2(z) = \frac{1}{2}, \quad
\log_2(y) - \log_2(x) - \log_2(z) = \frac{1}{3}, \quad
\log_2(z) - \log_2(x) - \log_2(y) = \frac{1}{4}.
\]

Let \(a = \log_2(x),\ b = \log_2(y),\ c = \log_2(z)\). Then:
\begin{align}
a - b - c &= \frac{1}{2} \tag{1} \\
b - a - c &= \frac{1}{3} \tag{2} \\
c - a - b &= \frac{1}{4} \tag{3}
\end{align}

Adding (1), (2), and (3), the LHS becomes:
\[
a - b - c + b - a - c + c - a - b = -a - b - c
\]
and the RHS:
\[
\frac{1}{2} + \frac{1}{3} + \frac{1}{4} = \frac{13}{12}
\Rightarrow a + b + c = -\frac{13}{12}
\]

Now solve for each:
- From (1): \(2a = \frac{1}{2} - \frac{13}{12} = -\frac{7}{12} \Rightarrow a = -\frac{7}{24}\)

- From (2): \(2b = \frac{1}{3} - \frac{13}{12} = -\frac{3}{4} \Rightarrow b = -\frac{3}{8}\)

- From (3): \(2c = \frac{1}{4} - \frac{13}{12} = -\frac{5}{6} \Rightarrow c = -\frac{5}{12}\)

Now compute:
\[
\log_2(x^4 y^3 z^2) = 4a + 3b + 2c
= 4\left(-\frac{7}{24}\right) + 3\left(-\frac{3}{8}\right) + 2\left(-\frac{5}{12}\right)
= -\frac{28}{24} - \frac{9}{8} - \frac{10}{12}
\]

Convert to common denominator 48:
\[
-\frac{56}{48} - \frac{54}{48} - \frac{40}{48} = -\frac{150}{48} = -\frac{25}{8}
\Rightarrow \left|\log_2(x^4 y^3 z^2)\right| = \frac{25}{8}
\Rightarrow \boxed{m+n = 25 + 8 = \boxed{33}}
\]

\vspace{4pt}
\noindent\rule{\linewidth}{0.4pt}
\vspace{6pt}

\textbf{Final Answer:} {\color{green!50!black}\(\boxed{33}\)}

\end{tcolorbox}

\end{document}